\documentclass{article}

\usepackage{spconf,amsmath,graphicx}
\usepackage{amsmath}
\usepackage{algorithm}
\usepackage[noend]{algpseudocode}
\usepackage{tabularx}
\usepackage{adjustbox}
\usepackage{float}

\makeatletter
\def\BState{\State\hskip-\ALG@thistlm}
\makeatother

\usepackage[belowskip=-5pt,aboveskip=0pt]{caption}
\usepackage{multicol}
\usepackage{multirow}

\title{Dataset Culling: Towards Efficient Training of Distillation-based Domain Specific Models}

\twoauthors
  {Kentaro Yoshioka}
	{Stanford University / Toshiba\\
	Computer Science \\
	kyoshioka47@gmail.com}
  {Edward Lee, Simon Wong, Mark Horowitz}
	{Stanford University \\
	Electrical Engineering and Computer Science \\
    {edhlee,sswong}@stanford.edu, horowitz@ee.stanford.edu}
	
\begin{document}

\maketitle

\vspace*{-1cm}
\begin{abstract}

Real-time CNN-based object detection models for applications like surveillance can achieve high accuracy but are computationally expensive. Recent works have shown $10$ to $100\times$ reduction in computation cost for inference by using domain-specific networks. However, prior works have focused on inference only. If the domain model requires frequent retraining, training costs can pose a significant bottleneck. To address this, we propose Dataset Culling: a pipeline to reduce the size of the dataset for training, based on the prediction difficulty. Images that are easy to classify are filtered out since they contribute little to improving the accuracy. The difficulty is measured using our proposed confidence loss metric with little computational overhead. Dataset Culling is extended to optimize the image resolution to further improve training and inference costs. We develop fixed-angle, long-duration video datasets across several domains, and we show that the dataset size can be culled by a factor of 300$\times$ to reduce the total training time by 47$\times$ with no accuracy loss or even with slight improvement. 
\footnote{Codes are available: https://github.com/kentaroy47/DatasetCulling}
\end{abstract}
% [fixed-angle video, confusion vs confidence]


%%%%%%%%%%%%%%%%%%%%%%%%%%%%%%%%%%%%%%%%%%%%%%%%%%%%%%%%%%%%%%%%%%%%%%%%%%%%%%%%%
\vspace*{-0.35cm}
\section{Introduction}
\vspace*{-0.25cm}

% Problem statement. maybe a bit long.
Convolutional neural network (CNN) object detectors have recently achieved significant improvements in accuracy \cite{ren2015faster}\cite{lin2018focal} but have also become more computationally expensive.
Since CNNs generally obtain better classification performance with larger networks, there exists a tradeoff between accuracy and computation cost (or efficiency). One way around this tradeoff is to leverage application and domain knowledge. For example, models for stationary surveillance and traffic cameras require pedestrians and cars to be detected but not different specifies of dogs. Therefore, by leveraging specialization, smaller models can be used. 

Recent approaches utilize domain-specialization to train compact domain specific models (DSMs) with distillation \cite{hinton2015distilling}. Compact student models can achieve high accuracies when trained with sufficient domain data, and such student models can be 10-100$\times$ smaller than the teacher. \cite{kang2017noscope} utilized this idea in a model cascade, \cite{mullapudi2018online} pushed this idea to the extreme by training frequently with extremely-small student models, and \cite{radosavovic2017data} used unlabeled data to augment the student dataset. 
%In our paper, we aim to improve recent state-of-art techniques to achieve high accuracy and computation efficiency in the context of surveillance. 

\begin{figure}[t]
\vspace*{-0.35cm}
\centering
  \includegraphics[width=0.45\textwidth]{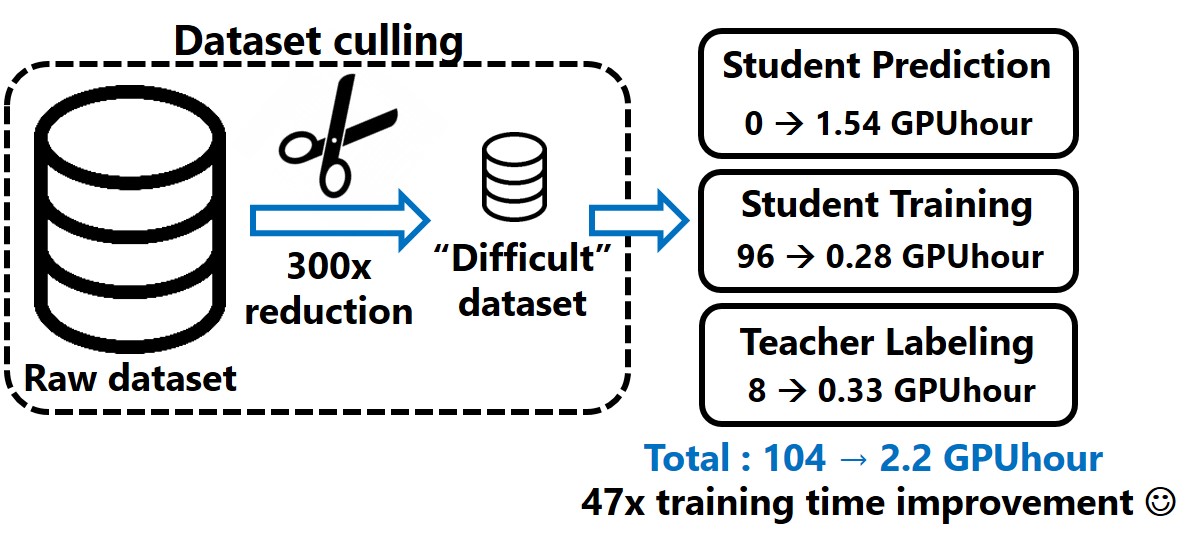}
  \caption{Dataset Culling aims to reduce the size of the unlabelled training data (number of images and image resolution) to reduce the computation costs for both the teacher and student. }
  \label{fig-overview}
  \vspace*{-0.35cm}
\end{figure}

The computation cost in conventional teacher-student frameworks is as follows: 1. Inference cost for the student, 2. Inference cost for the teacher (for labeling) and 3. Training cost for the student. Importantly, small student models may require frequent retraining to cancel out drift in data statistics associated with environment. For example, in a traffic surveillance setting, the appearance of pedestrian and cyclist may change seasonally. Hence, frequent retraining may be necessary when a small model is used, due to its capability to learn features is limited. Therefore with a small model, one can achieve computationally-efficient inference but with high (re)training overheads. For our surveillance application, a day's worth of surveillance data (86,400 images at 1 fps) requires over 100 GPU hours (Nvidia K80 on AWS P2) to train.

%%%%%%%%%%%%%%%%%%%%%%%%%%%%%
% figure of dataset culling technique

\begin{figure*}[ht!]

  \centering
  \includegraphics[width=1\textwidth]{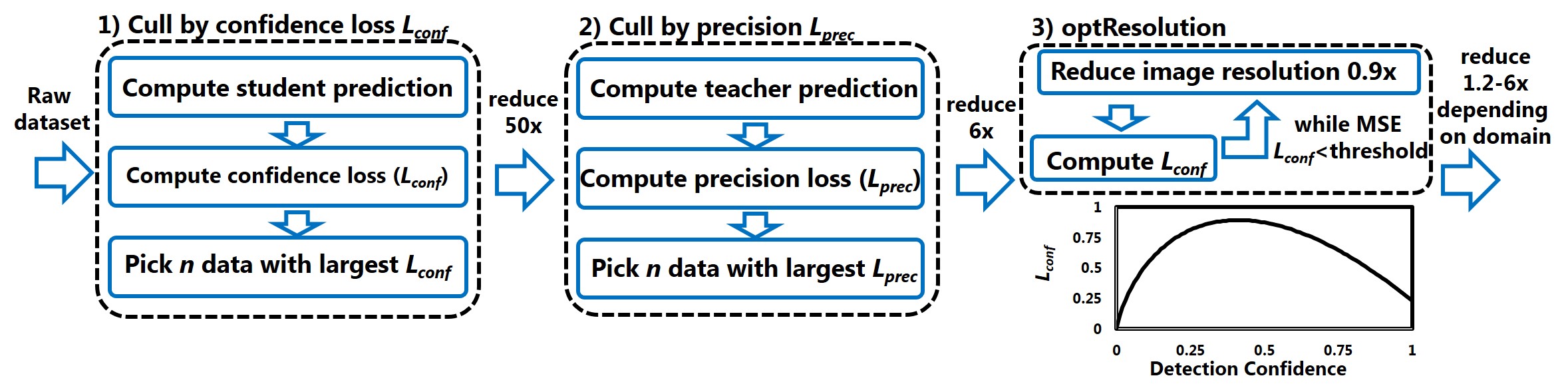}
  \caption{Our Dataset Culling pipeline. First, by culling the data with the confidence loss ($L_{conf}$), the dataset size is reduced 50$\times$ (in surveillance). The dataset is further reduced 6$\times$ by culling further with precision using teacher predictions. Finally, optResolution is applied to further reduce computation by another 1.2-6$\times$.}
  \label{fig-culling}
  \vspace*{-0.35cm}
\end{figure*}

%%%%%%%%%%%%%%%%%%%%%%%%%%%%%%%%%%

% Proposal
 Prior works have discussed ways to improve computation costs for the student model during inference. However, there has not been much focus on costs associated with (re)training or teacher costs. \cite{kang2017noscope} required the student and teacher to be run for all data samples for training, holding significant computing costs. Selection of important training data have been explored in active learning \cite{wang2017cnnactivelearning}\cite{settles2009active}, but less study has been done upon extremely reducing the {\it surveillance} dataset size.
 Our contributions are:
 %utilizing measures such as entropy but its goal was to mainly reduce the labeling efforts but not the total training time. We will show that  
% Contributions
\begin{itemize}
\vspace*{-0.25cm}
\setlength\itemsep{0em}
\item We propose Dataset Culling, which significantly reduces the computation cost of training. We show speedups in training by over 47$\times$ with no accuracy penalty. 
To the best of our knowledge, this is the first successful demonstration of improving training efficiency of DSMs.
\item  To achieve high accuracy and fast training, Dataset Culling operates in a pipeline: only student predictions are used to cull majority of the easy-to-predict data and precise filtering is done in the second stage using teacher predictions. The dataset size can be culled by a factor of 300$\times$.
\item We develop optResolution as part of the Dataset Culling pipeline. This  optimizes the image resolution of the dataset to further improve inference and training costs. 
\setlength\itemsep{0em}
\end{itemize}
\vspace*{-0.4cm}

%%%%%%%%%%%%%%%%%%%%%%%%%%%%%%%%%%%%%%%%%%%%%%%%%%%%%%%%%%%%%%%%%%%%%%%%%%%%%%
% figure of dataset + culling results.
% dataset is culled
%\vspace*{-0.5cm}
\begin{figure*}[ht!]
  \centering
  \includegraphics[width=0.85\textwidth]{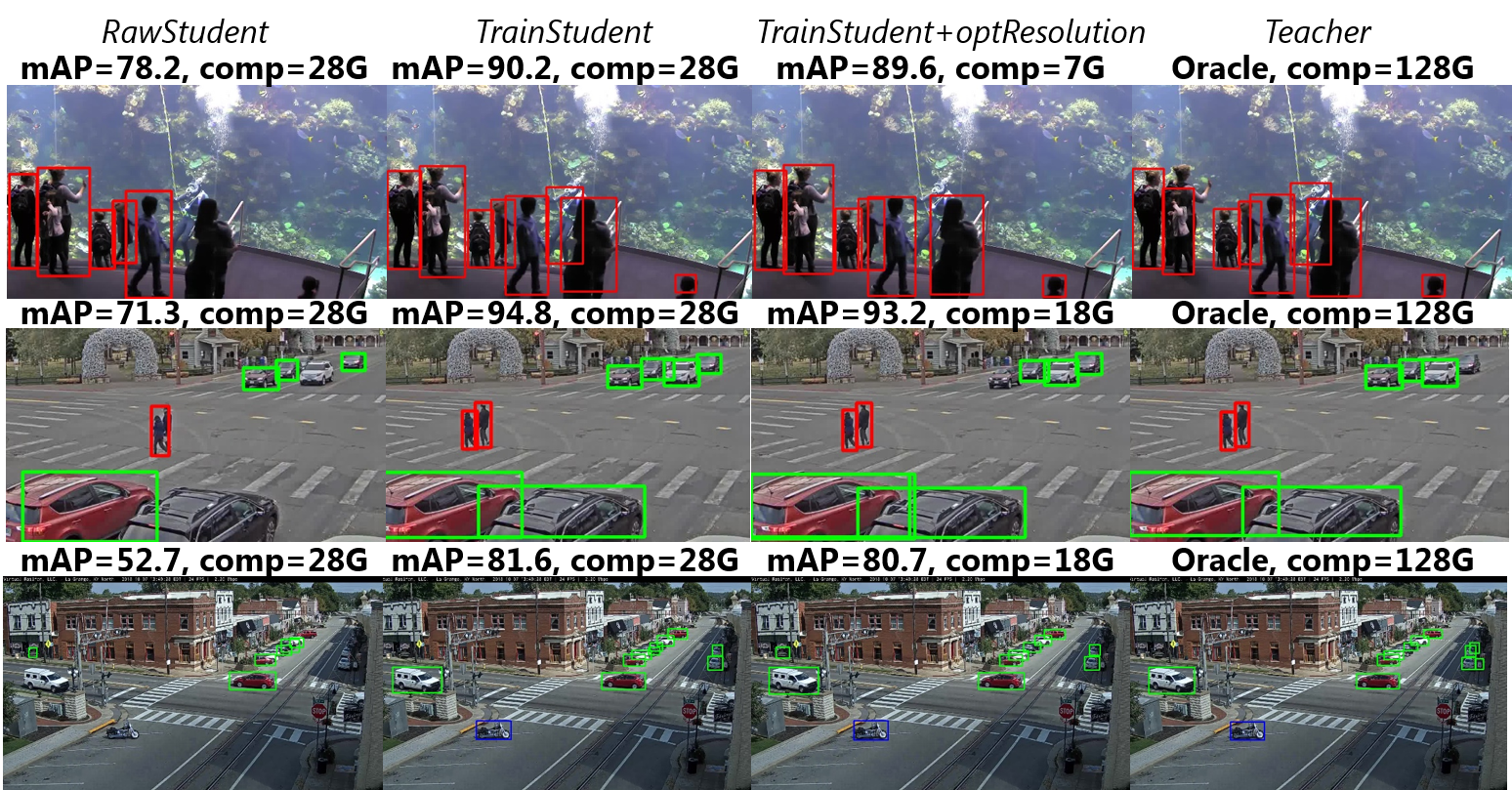}
  \caption{Object detection results of Dataset Culling with results of 3 scenes from surveillance dataset (top to bottom: Coral, Jackson, Kentucky.). Accuracy and the computation cost per image (GFLOPs) are shown. The student model is trained with a compressed dataset of $n$=128, and optResolution is set automatically. While optResolution introduces a penalty in accuracy (average 1\% mAP), it dramatically improves the computation cost. For example, the computation cost for inference is improved by up to 18$\times$ for Coral.}
  \label{fig-dataset}
  \vspace*{-0.25cm}
\end{figure*}
\vspace*{-0.35cm}
%%%%%%%%%%%%%%%%%%%%%%%%%%%%%%%%%%%%%%%%%%%%%%%%%%%%%%%%%%%%%%%%%

%% story of the main body

\section{Efficient training of DSMs}
\vspace*{-0.25cm}
The role of DSMs is to achieve high object detection performance with small model size. However, training of DSMs can itself be computationally problematic. In order to reduce training time, we propose Dataset Culling. This procedure removes (filters out) data samples from the training dataset that are believed to be easy to predict, and minimizes the image resolution of the dataset while maintaining accuracy (Fig. \ref{fig-overview}). By reducing both the dataset size and image resolution, we reduce the number of 1) expensive teacher inference passes for labelling, 2) the training steps of the student, and 3) computations for the student. 

\vspace*{-0.4cm}
\subsection{Dataset Culling.}
\vspace*{-0.1cm}
The Dataset Culling pipeline is illustrated in Fig.\ref{fig-culling}. We first assess the difficulty of a stream of data by performing inference through the student. During training, model parameters are only updated when differences between the label and prediction exist. In other words, "easy" data, which the student already provides good predictions for, do not contribute to training. The designed confidence loss (shown below) assesses the difficulty of prediction on a sample of data from the student's output probabilities. For example, if the model's output probability (i.e. confidence) for an object class is high, we assume that the sample of data is reasonably easy to infer, and similarly if the answer is very low, the region is likely to be background. However, intermediate values mean that the data is hard to infer.

To evaluate how difficult an image is to predict, we develop a confidence loss metric. This loss (shown below) uses the model's output confidence levels to determine whether data samples are 1) difficult-to-predict and kept or 2) easy and culled away. 
%\vspace*{-0.4cm}
\begin{eqnarray}
\label{eqn:loss}
\centering
  L_\mathrm{conf}  & = & -x\log{x} \times Q + (1 - x) \frac{\exp{x}}{\exp{x}+1} + b \nonumber
%\vspace*{-0.8cm}
\end{eqnarray}
Input $x$ is the prediction confidence, $b$ is a constant to set the intercept to zero, and $Q$ sets the weighting of low-confidence predictions. In experiments, we use $b=0.5$ and $Q=3$ to roughly weight low-confidence detections $3\times$ more than confident results. The absolute form of the loss function is not essential as we observed similar results by designing functions which emphasize unconfident predictions. When the model provides multiple detection results, $L_\mathrm{conf}$ is computed for each prediction and are summed to obtain the overall loss of the data. 
While the confidence loss is a similar measure to entropy \cite{shannon1948mathematical}, the loss takes a positive amount even with confidence = 1; the loss will become a function of the total number of objects in the image, to ensure that images without objects (e.g. images at midnight) are not misinterpreted as "difficult". Since object-less images contribute little to training, such images should be filtered. 
This first stage of culling yields a 10 to 50$\times$ reduction in the size of training data.

In the second stage of culling, we feed the remaining samples into the computationally-expensive teacher model. We compare the answers made between the teacher and student and use this to directly determine the difficulty. Here, we compute the average precision by treating the teacher predictions as ground truths. Using this second stage of culling, we further reduce the number of data samples by $6\times$ while sustaining the accuracy. Furthermore, in some cases, we can even improve the student's mAP as we eliminate data that add little to no feedback for enhancing the student. 

% image resolution changing
\vspace*{-0.3cm}
\subsection{Optimizing the image resolution (optResolution).}
\vspace*{-0.3cm}
Our second technique in Dataset Culling is optResolution, which sets the image resolution as a function of the prediction difficulty. By decreasing the CNN-input image size, we reduce the number of multiply-and-add operations and memory footprints. For example, with a $2\times$ reduction in image resolution, we obtain 4$\times$ improvement in computational efficiency. OptResolution takes advantage of the fact that object detection difficulties depend on the scene and application itself. For example, objects-of-interest in video for indoor and sport scenes are usually large in size and relatively easy to classify with low-resolution. However, traffic monitoring cameras demand high resolutions in order to monitor both relatively-small pedestrians and large vehicles. Traditionally, human supervision or expensive teacher inference was required to tune resolution \cite{jiang2018chameleon}. 

Dataset Culling integrates optResolution with low computational overheads. We first feed an image $x$ of (pixel) size $H\times W$ into the student model and compute the confidence loss and compare the confidence loss when the image is downsampled $x$ to size $0.9W \times 0.9H$. These downsampling operations are recursively performed until the confidence loss exceeds a predefined threshold, as it indicates that objects are becoming harder to infer. In our implementation, we compute the mean-squared-error (MSE) of the confidence loss against full-resolution inference results ($L\mathrm{conf}_{FR}$) as
\vspace*{-0.4cm}
\begin{eqnarray}
    \centering
    MSE & = &  \sum\limits_{i=0}^{N} (L\mathrm{conf}_{FR}[i] -  L\mathrm{conf}[i])^2 \nonumber
\end{eqnarray}
%\vspace*{-0.2cm}
Here, $L\mathrm{conf}$ is the confidence loss of the downsampled inference, and $N$ is the size of the culled dataset.
One limitation of optResolution is that we strongly assume that the overall object size is constant for both training and runtime. For example, pedestrians as viewed by a surveillance camera are assumed to remain roughly the same size from train-time to test-time unless the surveillance camera were to move to a different position or orientation. However, in such cases, the model would require retraining either way. 

% added limitations [ken]

%%%%%%%%

\begin{table*}[!th]
\centering
\caption{We evaluate how culling the dataset impacts accuracy. Here, we reduce the dataset size using the first two stages in Fig. 2. (cull by confidence and precision) without optResolution. Time is reported in GPU hours.}
\label{big-table-results}
\begin{tabular}{|c|c|c|c|c|c|c|c|}
\hline
\multirow{2}{*}{Dataset}      & \multirow{2}{*}{\begin{tabular}[c]{@{}c@{}}Training\\ images\end{tabular}} & \multirow{2}{*}{}                                                                                  & \multicolumn{5}{c|}{Target dataset size}                                                                                                                                                                                                              \\ \cline{4-8} 
                              &                                                                            &                                                                                                    & 64                                                         & 128                                                        & 256                                                        & Full                                                & No train \\ \hline
\multirow{3}{*}{Surveillance} & \multirow{3}{*}{86,400}                                                    & Accuracy {[}mAP{]}                                                                                 & \textbf{85.56 (- 3.0\%)}                                   & \textbf{88.3 (- 0.3\%)}                                    & \textbf{89.3 (+ 0.8\%)}                                    & 88.5                                                & 58.6     \\ \cline{3-8} 
                              &                                                                            & Total Train Time                                                                                   & \textbf{1.9 (54$\times$)}                                  & \textbf{2.0 (50$\times$)}                                  & \textbf{2.2 (47$\times$)}                                  & 104                                                 & -        \\ \cline{3-8} 
                              &                                                                            & \begin{tabular}[c]{@{}c@{}}Student Training\\ Student Prediction\\ Teacher Prediction\end{tabular} & \begin{tabular}[c]{@{}c@{}}0.07\\ 1.54\\ 0.33\end{tabular} & \begin{tabular}[c]{@{}c@{}}0.14\\ 1.54\\ 0.33\end{tabular} & \begin{tabular}[c]{@{}c@{}}0.28\\ 1.54\\ 0.33\end{tabular} & \begin{tabular}[c]{@{}c@{}}96\\ 0\\ 8\end{tabular}  & -        \\ \hline
\multirow{3}{*}{Sports}       & \multirow{3}{*}{3,600}                                                     & Accuracy {[}mAP{]}                                                                                 & \textbf{93.7 (- 0.1\%)}                                    & \textbf{93.8 (0\%)}                                        & \textbf{93.8 (0\%)}                                        & 93.8                                                & 80.7     \\ \cline{3-8} 
                              &                                                                            & Total Train Time                                                                                   & \textbf{0.16 (16$\times$)}                                 & \textbf{0.23 (11$\times$)}                                 & \textbf{0.40 (6$\times$)}                                  & 2.5                                                 & -        \\ \cline{3-8} 
                              &                                                                            & \begin{tabular}[c]{@{}c@{}}Student Training\\ Student Prediction\\ Teacher Prediction\end{tabular} & \begin{tabular}[c]{@{}c@{}}0.07\\ 0.06\\ 0.03\end{tabular} & \begin{tabular}[c]{@{}c@{}}0.14\\ 0.06\\ 0.03\end{tabular} & \begin{tabular}[c]{@{}c@{}}0.28\\ 0.06\\ 0.06\end{tabular} & \begin{tabular}[c]{@{}c@{}}2\\ 0\\ 0.5\end{tabular} & -        \\ \hline
\end{tabular}
\end{table*}
%%%%%%%%%%%%%%%%%%%%%%%%%%%%%%%%%%%%%%%%%%%%%%%%%%%%%%%%%%%%%%%%%%%%%%%
\vspace*{-0.5cm}
\section{Experiments}
\vspace*{-0.1cm}

\begin{table*}[th!]
\centering
\caption{Ablation study and comparison of Dataset Culling strategies conducted on Jackson dataset. Our approach of conducting both filtering by confidence loss and data difficulty has a good balance of accuracy and computation. Confidence-only culling misses more than $50\%$ of samples that was otherwise kept using Precision-only, and Confidence$+$Precision misses $20\%$ that Precision-only kept. All strategies have a target dataset size of 128.}
\label{ablation}
\begin{tabular}{|c|c|c|c|c|c|c|}
\hline
Filtering strategy & Intermittent Samp. & Entropy\cite{shannon1948mathematical} & Confidence & Precision & \textbf{Confidence + Precision} & Full dataset \\ \hline
mAP                & 0.731              & 0.866        & 0.911           & 0.954          & \textbf{0.948}                  & 0.958        \\ \hline
GPU hours          & 0.15               & 1.7          & 1.7             & 8.0            & \textbf{2.0}                    & 104          \\ \hline
\end{tabular}
\vspace*{-0.4cm}
\end{table*}
\vspace*{-0.25cm}
%%%%%%%%%%%%%%%%%

\textbf{Models. } 
For experiments, we develop Faster-RCNN object detection models pretrained on MS-COCO \cite{ren2015faster}\cite{lin2014microsoft}. We utilize resnet101 (res101) for the teacher and resnet18 (res18) for the student model for the region proposal network (RPN) backbone \cite{he2016deep}. We expect similar outcomes when MobileNet \cite{howard2017mobilenets} is utilized for RPN, since both models achieve similar Imagenet accuracy. We chose Faster-RCNN for its accuracy and adaptive image resolution, but Dataset Culling can be applied to other object detection frameworks such as SSD and YOLO with fixed resolutions, which have similar training frameworks \cite{redmon2016you}\cite{liu2016ssd}.

\textbf{Custom Long-Duration Dataset. } 
We develop 8 long-duration, fixed-angle videos from \texttt{YouTubeLive} to evaluate Dataset Culling. As manually labelling all frames in video is cumbersome, we label the dataset by treating the teacher predictions as ground truth labels as in \cite{kang2017noscope}. In this paper, we report the accuracy as mean average precision (mAP) at 50\% IoU.
We will cite 5 of the videos "surveillance", each consisting of 30-hour fixed-angle streams with  1 to 4 detection classes. The first 24 hours (86,400 images) are used for training and the consequent 6 hours (21,600 images) are used for validation.
We cite 3 of the videos "sports", which consist of 2-hours (7,200 images) of fixed-angle videos. Here, the class is "person" only and the training and testing images are split evenly.

\textbf{Results. }
Object detection results from 3 scenes are shown in Fig.\ref{fig-dataset} and mAP and computation costs are reported in Table \ref{big-table-results}. For surveillance results, using domain specific training for the student improved accuracy by 31\% compared to the COCO-pretrained student results. In comparison to the full dataset training, Dataset Culling improves the training time 47$\times$ by reducing the dataset size to $n$=256. A slight increase in accuracy is observed because Dataset Culling presents a similar effect to hard example mining \cite{shrivastava2016onlineharddatamining}, where training on a limited but difficult data benefits model accuracy. Since we include the time to run inference on the entire training set in the training time, further culling of the dataset does not dramatically improve the training time. However in smaller datasets (sports), increasing the culling ratio contributes to training efficiency because the model training time is a large fraction of training.

%%%%%%%%%%%%%%%%

\begin{figure}[th!]
\centering
  \includegraphics[width=0.45\textwidth]{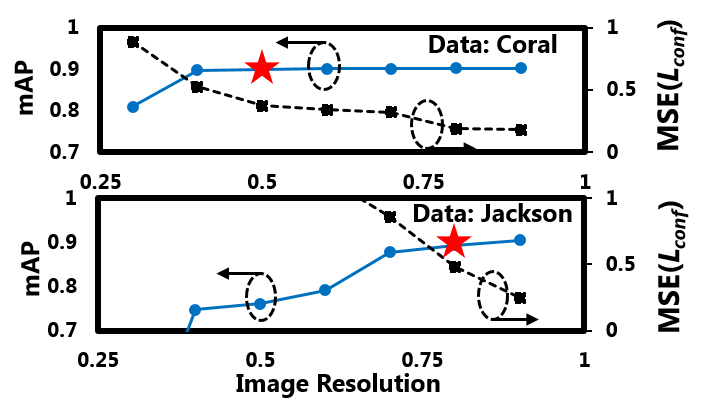}
  \caption{The image resolution was scaled manually to observe the change in mAP (blue solid line) and computed MSE of $L\mathrm{conf}$ (black dasheds) in two domains. The red star indicates the image resolution proposed by optResolution, meeting both high accuracy and computation cost.}
  \label{resolution-results}
  \vspace*{-0.3cm}
\end{figure}

%%%%%%%%%%%%%%%%

% ablation study.
\textbf{Ablation study. }
We perform ablation as shown in Table \ref{ablation}.
We construct a dataset (of size $n=128$) with only difficult-to-predict images with four filtering techniques and compare the mAPs of our trained student models. We show that while filtering using only the precision metric (no confidence-loss induced or image scaling) can achieve the highest accuracy, the training time is 4$\times$ higher than our final approach (confidence + precision). Finally, we illustrate that with both culling procedures, we can realize a good balance of accuracy and computation.

\textbf{optResolution.}
Fig. \ref{resolution-results} shows the results of optResolution. OptResolution provides a well-tuned image resolution, satisfying both accuracy and computation costs. For indoor surveillance (Coral), the objects-of-interest were large and easy to detect. OptResolution thus selects a resolution of 0.5$\times$. For surveillance of traffic in Jackson, the objects are small in size and difficult to detect. Therefore, our procedure selects a scale of 0.8$\times$. 

\vspace*{-0.45cm}
\section{Conclusions}
\vspace*{-0.3cm}
DSMs dramatically decrease the cost of inference. But if models need frequent retraining, training costs can become a significant bottleneck. We show how Dataset Culling significantly reduces training time and overall computation costs by 47$\times$ with little to no accuracy penalty in our long-duration, fixed-angle datasets.

% -------------------------------------------------------------------------
\bibliographystyle{IEEEbib}
\bibliography{main}

\end{document}